\documentclass[11pt]{article}

\usepackage[preprint]{acl}

\usepackage{times}
\usepackage{latexsym}

\usepackage[T1]{fontenc}

\usepackage[utf8]{inputenc}

\usepackage{microtype}

\usepackage{inconsolata}

\usepackage{graphicx}
\usepackage{booktabs}
\usepackage{multirow}

\usepackage{tikz}
\usetikzlibrary{trees, arrows.meta, positioning, calc}
\usepackage{enumitem}

\title{Building a Custom Taxonomy of AI Skills and Tasks\\ from the Ground Up with Job Postings}

\author{Stephen Meisenbacher \\
  Technical University of Munich \\
  Garching, Germany \\
  \texttt{stephen.meisenbacher@tum.de} \\\And
  Peter Norlander \\
  Loyola University Chicago \\
  Chicago, IL, USA \\
  \texttt{pnorlander@luc.edu} \\}

\begin{document}
\maketitle
\begin{abstract}
Utilizing LLMs for automated taxonomy construction presents a clear opportunity for the comprehensive, yet efficient mapping of potentially complex domains. When contending with high volumes of rapidly growing corpora, however, it becomes unclear how to best leverage such data for optimal taxonomy construction. Taking the case of systematizing \textit{AI skills in the workplace}, we use two large-scale job postings corpora to investigate key design decisions for the inclusion (or exclusion) of data points for taxonomy construction. We propose \textsc{TaxonomyBuilder} as a blueprint for our systematic study, with which we evaluate various configurations of custom, data-informed, and hierarchical taxonomies. We demonstrate that \textit{less} data can provide more clarity:  filtering inputs to \textsc{TaxonomyBuilder} provides better domain-specific coverage than offering unfiltered inputs to clustering and LLM-enhanced hierarchical taxonomy labeling tools.
\end{abstract}

\section{Introduction}
\textit{What AI Skills are in demand in recent years}? To answer this and other timely questions about rapid change in labor markets, novel, open taxonomies of tasks and skills are needed to enable tracking of granular changes in work \citep{frank_toward_2019,margaryan_artificial_2023}. When built from high-volume, real-time sources of labor market text such as job postings, such taxonomies are valuable sources of data to inform individual career development, educational program designs, workforce development interventions, and firm strategy \cite{zweig_job_2026}.

However, proprietary taxonomies, data access restrictions, and lack of standardization limit the development and use of automated tools to harness general information from job postings at scale \cite{national_academies_of_sciences_artificial_2025}. Open data, including novel taxonomies and data generated by them, are potential public goods to support broad skills development, training needs for workforce development, growth, democratization, and entrepreneurship \cite{lerner_simple_2002,nagaraj_improving_2020,nagaraj_private_2022}. From a research standpoint, to assess whether AI is complementing or replacing human skills \cite{acemoglu_building_2026}, social scientists need detailed and up-to-date taxonomic information on the human tasks and skills of workers building and using AI.

When faced with such problems, researchers have often turned to \textit{taxonomy construction}. The creation of taxonomies has spanned the decades, and established methodologies are both accessible and widely used \cite{michalski1983learning,nickerson2013method}. The shortcomings of taxonomy creation mirror those of many traditionally manual tasks, such as the expense of large-scale manual labor, researcher fatigue and bias, and the difficulty in empirically validating created artifacts with real-world data \cite{vu2025automated}. The creation of representative taxonomies in custom, complex, and domain-specific fields remains challenging.

In the field of Natural Language Processing and especially in the era of LLMs, researchers have begun to explore the capabilities of automating the process of taxonomy construction. The promise of such approaches is the ability to sift through and categorize massive amounts of data, a step that would be highly expensive in the traditional manner of hand-crafted taxonomies \cite{vu2025automated}. Improving initial approaches leveraging clustering algorithms \cite{gordon11996hierarchical,10.5555/3000001.3000093,ienco2008towards}, recent state-of-the-art methods have leveraged LLMs and their generative capabilities \cite{moskvoretskii2026large}. 

As a motivation for our work, when confronted with the need to develop a taxonomy of AI skills from job postings, the literature offers prompt engineering guidance \cite{10350471,vu2025automated} and handles situations where \textit{seed} information, or \textit{seed taxonomies} can guide the LLM-assisted taxonomy building process \cite{zhang-etal-2025-llmtaxo,motamedi2026semi}. However, in the situation at present, questions posed to us include: \textit{which data} should be used (among all data from a massive text corpus), \textit{how much} of this data should be used, and \textit{in what way} can the built taxonomies be evaluated in turn on real-world datasets? Thus, while it has been established that leveraging LLMs can be helpful for taxonomy building, there is a need for systematic experimentation with key design parameters for best practices in custom domains. 

To guide our experiments, we leverage two large corpora of job postings in the United States, spanning the past decade. To address the needs identified above, we build taxonomies that efficiently and comprehensively organize AI in the workplace, particularly skills, tasks, experience, and qualifications. To accomplish this, we design a simple, yet powerful pipeline that transforms mined candidate contexts into hierarchical taxonomies. We evaluate 12 configurations of our pipeline on automatic and LLM-based metrics, leading to clear recommendations for automatic taxonomy building.

Our findings reveal that when handling massive data corpora for automated taxonomy construction, \textit{less is more}. Although the tendency to include as much data as possible, including via data augmentation, might plausibly increase domain coverage, our results demonstrate the opposite. As such, we learn that careful augmentation, strict clustering, and selective filtering can contribute to comprehensive, yet concise taxonomies for large, data-rich domains such as AI skills in the workplace. Our work makes the following additional contributions:

\begin{enumerate}
    \itemsep 0em
    \item We deepen the capabilities of LLM-based taxonomy building by asking and answering key questions surrounding the setup and execution of automatic taxonomy creation pipelines.
    \item We showcase the ability of our proposed pipeline, \textsc{TaxonomyBuilder}, to efficiently and comprehensively map complex domains such as \textit{AI Skills in the Workplace}, replicated on two large-scale job posting corpora.
    \item We make the resulting AI skills taxonomies publicly available and open for further use.
    \item We open-source \textsc{TaxonomyBuilder} as a generalizable tool to any domain of interest, available at \url{https://github.com/sjmeis/TaxonomyBuilder} and as a Python package (\href{https://pypi.org/project/TaxonomyBuilder/}{\textsc{TaxonomyBuilder}}).
\end{enumerate}

\section{Related Work}
\paragraph{Automated taxonomy construction.}
Automated taxonomy construction, often discussed alongside \textit{ontology construction}, \textit{ontology engineering}, or \textit{taxonomy induction}, has been approached in diverse ways even before the era of LLMs.  Early methods focus on lexico-syntactic patterns to define relationships in a hierarchy \cite{hearst-1992-automatic}, which are efficient yet suffer from low recall. Subsequent methods relied on the distributional hypothesis \cite{pado-lapata-2007-dependency}, using the backbone of clustering algorithms and early text representations (e.g., bag of words). Hierarchical approaches \cite{gordon11996hierarchical, fountain-lapata-2012-taxonomy}, allow for the vertical scaling of taxonomies, extending beyond similarity-based clustering of related terms. Other approaches supplement these methods with rich existing knowledge bases such as WordNet \cite{10.1145/1031171.1031194,kozareva-hovy-2010-semi}.

Automated taxonomy construction has been aided by probabilistic and model-based approaches \cite{snow-etal-2006-semantic,poon-domingos-2010-unsupervised} or graph-based methods \cite{velardi-etal-2013-ontolearn}. Building on the success of embeddings as text representations, ensuing works leveraged these embeddings to form hybrid taxonomy construction methods \cite{fu-etal-2014-learning,espinosa-anke-etal-2016-supervised,10.1145/3219819.3220064}. Other works utilized modern reinforcement learning \cite{mao-etal-2018-end} or transfer learning \cite{navarro2020automated}.

\begin{figure*}[t!]
    \centering
    \includegraphics[width=\linewidth]{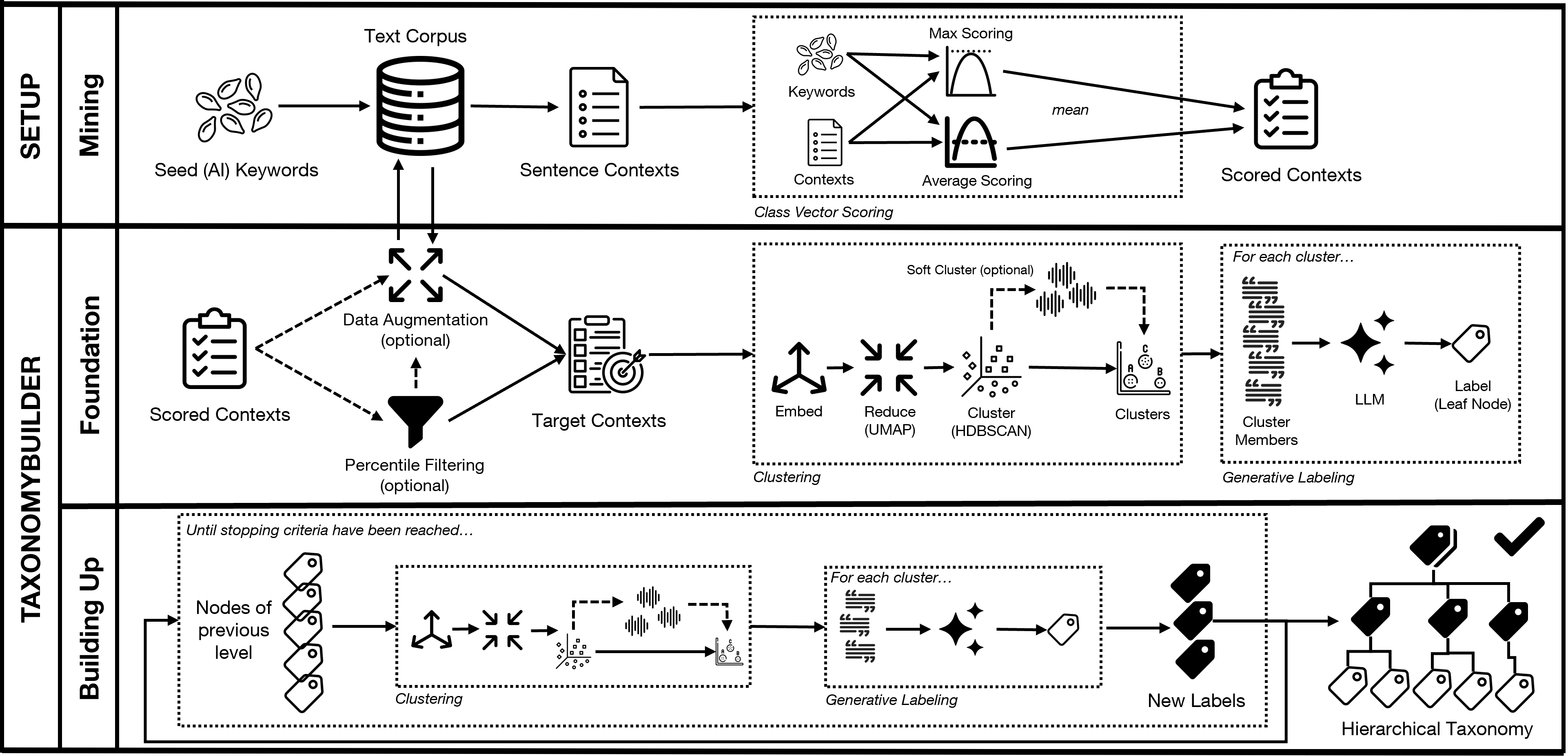}
    \caption{The \textsc{TaxonomyBuilder} method. In the top lane, we detail the setup method we follow as a precursor to taxonomy construction, which consists of keyword-based context mining and class-based scoring. The \textsc{TaxonomyBuilder} method, in turn, consists of two primary stages (depicted in the center and bottom lanes): (1) the construction of the foundation (leaf) level, followed by iterative vertical construction of further levels of the taxonomy hierarchy. We refer the readers to the main body for specific details on all stages of the process.}
    \label{fig:taxonomybuilder}
\end{figure*}

\paragraph{LLM-assisted taxonomy construction.}
The emergence of LLMs has sparked research interest in improving automated taxonomy construction \cite{vu2025automated}. Many approaches have been proposed in recent years, including methods to augment existing taxonomies \cite{kargupta-etal-2025-taxoadapt} as well as those that specialize in specific domains, such as scholarly research \cite{zhu-etal-2025-context,lahiri-etal-2025-taxoalign}. Typical LLM-based workflows include an extraction and clustering phase \cite{10.1145/3767695.3769519}, followed by generative labeling via prompting or fine-tuning \cite{10350471}, and finally taxonomy construction \cite{li-etal-2025-building,motamedi2026semi}. While this workflow is bottom-up (starting from leaf nodes), others follow a top-down approach \cite{10.1145/3627673.3679608,marchenko2024taxorankconstruct,10.1145/3637528.3671647}.

Our work adopts the bottom-up approach, as we investigate the case of modeling AI skills in job postings, for which no defined taxonomies exist. We build upon the work of approaches such as \textsc{LLMTaxo} \cite{zhang-etal-2025-llmtaxo}, who lay the groundwork for cluster labeling methods with LLMs (as well as LLM-based evaluation), and \textsc{TaxoAdapt} \cite{kargupta-etal-2025-taxoadapt}, which introduce important inter-mechanism redundancy checks. In terms of domain, the closest work to date is that of \citet{li-etal-2025-building}, who also operate in the domain of job posting data. However, we note that \citet{li-etal-2025-building} explore LLM-assisted taxonomy construction on very limited data samples (with a maximum of just over 10k job postings), whereas we study a case with millions to tens of millions of unique data points in question. In this, we introduce and systematically evaluate important questions surrounding optimal data usage, which has been previously unexplored, and evaluation, which becomes highly relevant in big data scenarios to ensure proper domain coverage.

\section{Datasets}
\label{sec:datasets}
We use two large corpora of job postings from the United States. The first is a collection of over 30 million job postings from the National Labor Exchange (NLx) between 2024 and 2025. To maintain reproducibility, we also evaluate our proposed method on an open corpus of 1.5 million federal job postings collected by \citet{Resh2025}. Both datasets are described briefly below.

\paragraph{NLx.}
The National Labor Exchange (NLx) ``provides workforce development professionals, academic researchers, employers, and other organizations that rely on labor market information (LMI) with high-quality, transparent, real-time and historical data that represents the diversity of jobs available in the labor market''\footnote{\url{https://nlxresearchhub.org/about-us/}} As of March 2026, the exchange hosts nearly 3.1 million open positions from 300,000 employers.

The entirety of the NLx corpus currently comprises over 173 million unique job postings from 2007 onwards. To maintain a manageable subset and to focus in on years which feature modern AI skill requirements, we narrow down our dataset to data from 2024 and 2025. Together, this results in a dataset of 32.12 million job postings within the two-year timespan. The average word count of the job postings is 875 (using \textsc{nltk} on a sample of all December 2025 postings). We make no modifications to the original NLx job posting texts.

Use of NLx data precludes sharing any data in its original form. As such, we replicate our experiments on a second, publicly available dataset.

\paragraph{USAJOBS.}
We also utilize a corpus of 1.54 million job postings from the \href{https://www.usajobs.gov/}{USAJOBS} platform, the US government's official employment website, made available by \citet{Resh2025}. This corpus contains job postings for positions in the federal government from 2017-2023. The USAJOBS postings are generally long, with an average word count of 5054 (as per \textsc{nltk} on November 2023 postings).

\paragraph{Test set.}
It should be noted that we reserve the last month's data in each corpus for evaluation purposes, i.e., December 2025 of NLx and November 2023 of USAJOBS. Thus, all procedures described in the next section, in which data is mined or augmented, leave out this reserved test set.

\paragraph{Data release.}
We make the USAJOBS dataset publicly available\footnote{\url{https://huggingface.co/datasets/loyoladatamining/usajobs} Note: the corpus has been updated to include more postings than were used in this work.}. Regarding outputs and analysis of NLx results, we cannot share the original data, a note we expand upon in our Ethics Statement.

\section{The \textsc{TaxonomyBuilder} Method}
\label{sec:method}
In the following, we detail the design and implementation of \textsc{TaxonomyBuilder}, a simple, lightweight, and efficient library for scaling from large corpora of unstructured texts to comprehensive and hierarchical taxonomies of concepts. The entire method is graphically visualized in Figure \ref{fig:taxonomybuilder}.

\subsection{The Setup: Mining contexts of interest}
The first stage of taxonomy construction, which we formally placed outside of the confines of \textsc{TaxonomyBuilder}, is the curation of candidate \textit{contexts} (i.e., sentences) which will serve as the basis for an AI skills taxonomy. Using our large corpus of job postings, we take a keyword-based approach for an initial efficient mining of candidates, followed by a scoring procedure via \textit{class vectors}.

\paragraph{Candidate mining.}
Due to very large-scale nature of our text corpora, we opt for an efficient, coarse-grained keyword-based search for the initial extraction of candidate contexts for an AI skills taxonomy. To build a comprehensive list of AI-related keywords, we consult various previous works that curate dictionaries of AI keywords sources \cite{baruffaldi_identifying_2020, alekseeva_demand_2021,lou_ai_2021,goldfarb_could_2023,lightcast2023openskills,maslej_artificial_2025,tambe_reskilling_2026}.

We compiled the keywords from the above sources, performed light editing (e.g., splitting entries with `/' into two and removing  parentheses), and then de-duplicated the list. This yielded a final set of 590 AI keywords, found in our repository.

We then performed an efficient keyword search using the \textsc{pyahocorasick} library, keeping all sentences matching one or more of the keywords, with either a white space or sentence boundary (to avoid inner-string matches). We also enumerated the number of sentence matches per job posting. This keyword-based search yielded 5.58 million candidate sentences for NLx and 213k for USAJOBS. As a further filtering step, we kept only the candidate sentences from job postings with three or more matches (so as to remove noise and non-AI jobs). This resulted in a final set of 251k candidates for NLx and 19k candidates for USAJOBS.

\paragraph{Candidate scoring.}
Recognizing that a keyword-search introduces the potential for false positives (a prime example being \textit{torch} as a welding instrument and not the machine learning library), we leveraged an efficient yet powerful method to score and rank all extracted candidates, allowing for further filtering. Motivated by the idea of \textit{class label vector representations} \cite{schopf2020semantic} and mean-max class scoring \cite{meisenbacher-etal-2024-improved}, we devised a hybrid scoring method that compares all candidate sentences to all keywords (here, our AI keywords) via cosine similarity of their sentence embeddings \cite{reimers-gurevych-2019-sentence}. 

For each given candidate sentence, its embedding is compared to all keyword embeddings, and then mean and max similarity scores are averaged to produce a final class-relatedness score. For numerical stability, we calculate this score on the basis of three embedding models: \textsc{all-MiniLM-L12-v2} \cite{reimers-gurevych-2019-sentence}, \textsc{gte-large} \cite{li2023generaltextembeddingsmultistage}, and \textsc{embeddinggemma-300m} \cite{vera2025embeddinggemmapowerfullightweighttext}. The scores of these three models were averaged for the final score.

We note that while the three model approach was used for this work, the released \textsc{TaxonomyBuilder} code currently only supports class-based filtering with one embedding model of choice.

\begin{figure*}[t]
    \centering
    \resizebox{0.9\textwidth}{!}{%
        \begin{tikzpicture}[
            box/.style={
                draw=cyan!70!black, thick, rounded corners, fill=cyan!5,
                align=center, font=\sffamily\scriptsize, inner sep=5pt
            },
            leaf/.style={
                box, fill=gray!10, draw=gray, text width=5.8cm, 
                align=left, font=\sffamily\tiny
            },
            level1/.style={box, fill=teal!10, text width=3.2cm, font=\sffamily\bfseries\scriptsize},
            level2/.style={box, text width=3.5cm},
            level3/.style={box, text width=3.5cm},
            line/.style={draw=gray, thick, -Latex, rounded corners}
        ]

        \node[leaf] (L4_1) {\textit{Example:} "Experience in performance analysis, modeling, and optimization of machine learning systems..."\\[3pt]};
        
        \node[leaf, below=0.3cm of L4_1] (L4_2) {\textit{Example:} "Proficient in developing and applying various numerical algorithms and optimization techniques..."\\[3pt]};
        
        \node[leaf, below=0.6cm of L4_2] (L4_3) {\textit{Example:} "Develop state-of-the-art machine learning models and algorithms to solve complex real-world problems..."\\[3pt]};
        
        \node[leaf, below=0.3cm of L4_3] (L4_4) {\textit{Example:} "Possesses hands-on understanding and experience with deep learning and machine learning algorithms..."\\[3pt]};

        \node[level3, left=0.8cm of L4_1] (L3_1) {Experience in performance analysis and modeling};
        \node[level3, left=0.8cm of L4_2] (L3_2) {Apply advanced numerical and ML algorithms};
        \node[level3, left=0.8cm of L4_3] (L3_3) {Develop advanced ML solutions for real-world applications};
        \node[level3, left=0.8cm of L4_4] (L3_4) {Apply machine learning techniques to real-world problems};

        \node[level2, left=0.8cm of L3_1] (L2_1) {Optimize machine learning and deep learning systems};
        \node[level2, left=0.8cm of L3_2] (L2_2) {Develop and apply advanced machine learning models};

        \path (L3_3.west) -- (L3_4.west) coordinate[midway] (mid_L3_bottom);
        \node[level2, left=0.8cm of mid_L3_bottom, anchor=east] (L2_3) {Apply ML and AI solutions to real-world challenges};

        \path (L2_1.west) -- (L2_2.west) coordinate[midway] (mid_L2_top);
        \node[level1, left=0.8cm of mid_L2_top, anchor=east] (L1_1) {Machine Learning Model\\Development \& Optimization};
        
        \node[level1, left=0.8cm of L2_3] (L1_2) {AI \& Machine Learning\\Integration};

        \path (L1_1.west) -- (L1_2.west) coordinate[midway] (mid_L1);
        \node[box, fill=blue!10, font=\sffamily\bfseries\small, left=0.8cm of mid_L1, anchor=east] (Root) {AI Skills\\Taxonomy};

        \draw[line] (Root.east) -- ++(0.3,0) |- (L1_1.west);
        \draw[line] (Root.east) -- ++(0.3,0) |- (L1_2.west);

        \draw[line] (L1_1.east) -- ++(0.3,0) |- (L2_1.west);
        \draw[line] (L1_1.east) -- ++(0.3,0) |- (L2_2.west);
        \draw[line] (L1_2.east) -- ++(0.3,0) -- (L2_3.west);

        \draw[line] (L2_1.east) -- ++(0.3,0) -- (L3_1.west);
        \draw[line] (L2_2.east) -- ++(0.3,0) -- (L3_2.west);
        \draw[line] (L2_3.east) -- ++(0.3,0) |- (L3_3.west);
        \draw[line] (L2_3.east) -- ++(0.3,0) |- (L3_4.west);

        \draw[line] (L3_1.east) -- (L4_1.west);
        \draw[line] (L3_2.east) -- (L4_2.west);
        \draw[line] (L3_3.east) -- (L4_3.west);
        \draw[line] (L3_4.east) -- (L4_4.west);

        \end{tikzpicture}%
    }
    \caption{Abridged example of the taxonomy structure produced by \textsc{TaxonomyBuilder} for AI skills. The right-most nodes exemplify the initial foundation built from raw context candidates. Our methods then proceeds to build upwards (``left'') upon this foundation. Note that \textit{AI Skills Taxonomy} is inserted for completion.}
    \label{fig:taxonomy_example}
\end{figure*}
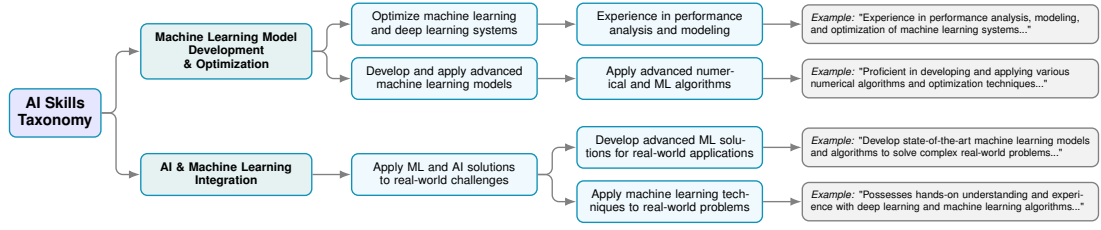

\subsection{Forming the foundation}
With the (large) set of candidate sentences mined, the next step becomes to distill these statements into a foundational layer of a taxonomy, one that is both representative of their diversity expressed among the candidates, but that also significantly compresses this information to form an accessible taxonomy. This process represents the first stage of \textsc{TaxonomyBuilder}: \textit{forming the foundation}.

\paragraph{Preliminaries: more and/or less data?}
How to effectively and intelligently leverage massive text corpora for an AI skills taxonomy using  \textsc{TaxonomyBuilder} requires investigation and evaluation. Given the core set of candidate statements described above, we explore two divergent paths, asking the question of whether \textit{more} (via augmentation) and/or \textit{less} (via filtering) data is most sensible.

The first option, \textit{data augmentation}, supplements the high-precision, low-recall, keyword-based candidate search with efficient embedding similarity augmentation. Given the set of (embedded) candidates, we perform a semantic search against all the documents in the respective corpus, \textit{augmenting} the candidate set with any new contexts found that exceed a cosine similarity threshold (we choose 0.9). For this procedure, we solely use \textsc{embeddinggemma-300m}. Using this threshold augmented the NLx candidates with an additional 746k candidates and 31k candidates for USAJOBS.

We also explore the option of \textit{less} data, in order to filter out noisy or irrelevant statements that have received a low class score. In this work, we experiment with keeping candidates (by score) above the top 75th, 50th, and 25th percentile. Note that such filtering also cascades to augmented candidates: if an original candidate is filtered out, all of its corresponding augmented sentences are as well.

\paragraph{Clustering.}
Given the target candidate contexts (following optional augmentation or filtering), the next step in \textsc{TaxonomyBuilder} is to cluster these contexts. Following the recommendations of \citet{zhang-etal-2025-llmtaxo}, we choose HDBSCAN as the clustering algorithm \cite{campello2013density}, using the GPU-accelerated library provided by rapids.ai\footnote{\url{https://rapids.ai/cuml-accel/}}. We perform clustering on embeddings that have been dimensionality reduced via UMAP to 10 dimensions (also using the \textsc{cuml} acceleration). We set the minimum cluster size to 5.

Since HDBSCAN by design assigns data points to a ``noise cluster'' if no fitting cluster is found, we explore the option of re-introducing such points to the clustering results via \textit{soft clustering}. This process involves calculating the distance of a noise point to all cluster centroids, and assigning it the cluster of closest proximity. We leave this option as a third point of investigation, along with data augmentation and percentile filtering.

The results of this process is a collection of clustered candidates, each of which represent a single entry in the bottom (leaf) level of the taxonomy.

\paragraph{Generative cluster labeling.}
To transform clusters of related statements into concise entries for a taxonomy foundation, we leverage the generative capabilities of LLMs to produce accurate representations (``labels'') of each cluster. This technique has been used in several previous works as an effective distillation method \cite{10350471,10.1145/3627673.3679608,zhang-etal-2025-llmtaxo}. For our main experiments, we use \textsc{gpt-4o-mini} \cite{openai2024gpt4ocard} with the prompt in Table \ref{tab:prompt_base} of the Appendix.

\paragraph{Label verification and pruning.}
To avoid error or redundancy propagation, we introduce a similarity-based check to the generated labels before proceeding to hierarchical taxonomy building. Specifically, the pairwise cosine similarity between each label embedding and all other labels is calculated, and all sets of labels exceeding a similarity threshold (here, 0.95) are aggregated into a consolidated label (using the zero-shot prompt of Table \ref{tab:prompt_agg}). The resulting collection of labels represent the leaf nodes of the taxonomy to be constructed.

\subsection{Building up}
The final stage of \textsc{TaxonomyBuilder} involves the iterative (vertical) creation of subsequent levels of the taxonomy. Given the foundational level labels, these are then used as the input for the next level, following the same embedding-clustering-labeling pipeline as before. This process continues until one of two stopping criteria are met: (1) the number of labels generated at a given label are less than a user-defined value (default: 10), or (2) the number of levels generated has reached a user-defined threshold (default: 5). This mitigates the possibility of either scarce top levels, or taxonomies that are too tall and therefore hard to interpret.

An example AI skills taxonomy, excerpted from an output of our experiments, is depicted in Figure \ref{fig:taxonomy_example}. The full taxonomies resulting from all experimental configurations are located in our repository.

\section{Experimental Setup}
The primary objective of our experiments is to evaluate the downstream effects of key design choices in the inclusion of data for taxonomy construction using \textsc{TaxonomyBuilder}. In particular, we study the impact of data augmentation, filtering measures, and soft clustering on the internal and external validity of the constructed AI skills taxonomies. Thus, we employ an intrinsic clustering-specific metric for clustering performance, as well as LLM-as-a-Judge criteria to evaluate the produced taxonomies as a whole. Finally, we empirically measure external validity by testing how well the constructed taxonomies generalize to real-world data, i.e., our held-out job posting test sets.

\subsection{Taxonomy configurations}
Using our two datasets, we construct an array of taxonomies based on different configurations, enabling a comparative study of key design decisions. In particular, we vary the following configurations:

\begin{itemize}[leftmargin=0.85em]
    \itemsep -0.3em
    \item \textbf{Data Augmentation}: whether the original candidate sets are augmented with new examples from the corpus. We utilize \textsc{embeddinggemma-300m} with a similarity threshold of 0.9.
    \item \textbf{Percentile Filtering}: leveraging class vector scoring, at which score threshold are candidates filtered out. We choose three configurations: top 25th, 50th, and 75th percentile of scores, i.e., only keeping candidates within these percentiles.
    \item \textbf{Soft Clustering}: whether to include noise points back in clusters or not (via centroid proximity).
\end{itemize}

With the above configurations, we create and evaluate \textbf{12} taxonomies for each dataset (24 in total), leveraging the \textit{TaxonomyBuilder} method.  

\subsection{Evaluation Metrics}
We select three classes of evaluation metrics, spanning \textit{clustering evaluation}, \textit{LLM-based taxonomy assessment}, and \textit{taxonomy domain coverage}.

\paragraph{Clustering evaluation.}
As the variation of our configurations, described above, can reasonably have a direct influence on the outcome of the HDBSCAN clustering, we employ a well-known cluster evaluation metric, namely the silhouette score. A higher scores in this metric implies that clusters are more well-defined. We use the \textsc{sklearn} implementation, and we average the per-layer scores across all levels of a constructed taxonomy. 

\begin{table*}[htbp]
\centering
\resizebox{0.99\textwidth}{!}{
\begin{tabular}{lcc c cccc cccc cccc c}
\toprule
\multirow{2}{*}{\textbf{Dataset}} & \multicolumn{2}{c}{\textbf{Configuration}} & \multirow{2}{*}{\textbf{Silh.}} & \multicolumn{4}{c}{\textbf{LLM-as-a-Judge (Category Avg.)}} & \multicolumn{4}{c}{\textbf{Lenient Coverage @ $\tau$}} & \multicolumn{4}{c}{\textbf{Strict Coverage @ $\tau$}} & \multirow{2}{*}{\textbf{Best Util.}} \\
\cmidrule(lr){2-3} \cmidrule(lr){5-8} \cmidrule(lr){9-12} \cmidrule(lr){13-16}
& \textbf{Aug. / Soft} & \textbf{Pct.} & & \textbf{Clar.} & \textbf{H.Coh.} & \textbf{Orth.} & \textbf{Comp.} & \textbf{0.9} & \textbf{0.8} & \textbf{0.7} & \textbf{0.6} & \textbf{0.9} & \textbf{0.8} & \textbf{0.7} & \textbf{0.6} & \\
\midrule
\multirow{12}{*}{\textbf{NLx}} 
& Y / Y & 25 & 0.047 & 2.62 & 3.08 & 1.62 & 3.02 & 0.472 & \textbf{0.595} & 0.585 & 0.346 & 0.498 & 0.545 & 0.447 & 0.227 & 0.580 \\
& Y / Y & 50 & 0.056 & 2.86 & 3.25 & 1.81 & 2.96 & 0.470 & 0.592 & 0.603 & 0.376 & 0.497 & 0.547 & 0.464 & 0.253 & 0.604 \\
& Y / Y & 75 & 0.092 & 3.07 & 3.57 & 2.21 & \textbf{3.29} & 0.462 & 0.579 & 0.630 & 0.427 & 0.494 & 0.557 & 0.494 & 0.297 & 0.610 \\
\cmidrule(lr){2-17}
& Y / N & 25 & -0.000 & 2.89 & 3.07 & 1.79 & 2.90 & \textbf{0.475} & 0.592 & 0.576 & 0.316 & \textbf{0.500} & 0.540 & 0.436 & 0.201 & 0.580 \\
& Y / N & 50 & 0.032 & 2.83 & 3.15 & 1.85 & 2.93 & 0.473 & 0.590 & 0.597 & 0.362 & 0.499 & 0.543 & 0.458 & 0.241 & 0.608 \\
& Y / N & 75 & 0.074 & 2.82 & 3.36 & 2.14 & 2.90 & 0.465 & 0.576 & 0.626 & 0.402 & 0.495 & 0.553 & 0.488 & 0.276 & 0.604 \\
\cmidrule(lr){2-17}
& N / Y & 25 & 0.073 & 2.75 & 3.20 & 1.90 & 3.07 & 0.460 & 0.554 & 0.702 & 0.546 & 0.493 & \textbf{0.600} & 0.599 & 0.398 & 0.468 \\
& N / Y & 50 & \textbf{0.111} & \textbf{3.25} & \textbf{3.75} & \textbf{2.25} & 3.22 & 0.457 & 0.538 & \textbf{0.707} & 0.602 & 0.489 & 0.598 & 0.636 & 0.445 & 0.814 \\
& N / Y & 75 & 0.065 & 2.88 & 3.25 & 2.00 & 3.17 & 0.456 & 0.504 & 0.682 & \textbf{0.663} & 0.489 & 0.570 & \textbf{0.672} & \textbf{0.505} & 0.844 \\
\cmidrule(lr){2-17}
& N / N & 25 & 0.026 & 2.92 & 3.33 & 1.83 & 2.89 & 0.461 & 0.558 & 0.700 & 0.533 & 0.497 & 0.596 & 0.589 & 0.385 & 0.447 \\
& N / N & 50 & 0.079 & 2.83 & 3.50 & 1.83 & 3.22 & 0.459 & 0.541 & 0.705 & 0.599 & 0.493 & 0.596 & 0.623 & 0.441 & 0.837 \\
& N / N & 75 & 0.096 & 2.92 & 3.50 & 2.00 & 3.11 & 0.456 & 0.510 & 0.685 & 0.648 & 0.489 & 0.576 & 0.662 & 0.488 & \textbf{0.851} \\
\midrule
\multirow{12}{*}{\textbf{USAJOBS}} 
& Y / Y & 25 & 0.155 & 3.34 & 2.70 & 2.14 & 2.73 & 0.485 & \textbf{0.514} & 0.543 & 0.319 & \textbf{0.510} & 0.518 & 0.469 & 0.244 & 0.294 \\
& Y / Y & 50 & 0.159 & 3.40 & 2.81 & 2.13 & 2.84 & 0.483 & 0.496 & 0.573 & 0.436 & 0.506 & 0.526 & 0.518 & 0.354 & 0.300 \\
& Y / Y & 75 & 0.430 & 2.75 & 1.75 & 1.50 & 2.33 & 0.481 & 0.490 & 0.586 & 0.584 & 0.501 & 0.525 & 0.559 & 0.466 & 0.823 \\
\cmidrule(lr){2-17}
& Y / N & 25 & 0.561 & 3.37 & 2.76 & 2.28 & 2.78 & \textbf{0.486} & \textbf{0.514} & 0.553 & 0.330 & 0.507 & 0.518 & 0.476 & 0.254 & 0.296 \\
& Y / N & 50 & 0.516 & 3.47 & 2.79 & 2.26 & 2.95 & 0.483 & 0.498 & 0.576 & 0.425 & 0.506 & 0.525 & 0.517 & 0.345 & 0.317 \\
& Y / N & 75 & 0.577 & 3.08 & 2.50 & 1.50 & 2.56 & 0.481 & 0.491 & 0.583 & 0.556 & 0.500 & 0.531 & 0.552 & 0.446 & 0.839 \\
\cmidrule(lr){2-17}
& N / Y & 25 & 0.345 & 3.33 & 1.61 & 1.56 & 2.04 & 0.483 & 0.492 & \textbf{0.587} & 0.587 & 0.505 & \textbf{0.534} & 0.572 & 0.469 & \textbf{0.872} \\
& N / Y & 50 & 0.362 & 3.50 & 1.93 & 2.07 & 2.05 & 0.481 & 0.489 & 0.577 & 0.607 & 0.503 & 0.527 & 0.588 & 0.486 & 0.862 \\
& N / Y & 75 & 0.112 & \textbf{3.62} & \textbf{3.30} & \textbf{2.46} & 3.06 & 0.480 & 0.484 & 0.542 & \textbf{0.644} & 0.499 & 0.516 & \textbf{0.637} & \textbf{0.526} & 0.829 \\
\cmidrule(lr){2-17}
& N / N & 25 & 0.673 & 3.33 & 1.71 & 1.67 & 1.97 & 0.482 & 0.491 & 0.586 & 0.589 & 0.506 & 0.533 & 0.571 & 0.470 & 0.835 \\
& N / N & 50 & \textbf{0.703} & 2.50 & 1.50 & 1.50 & 2.33 & 0.482 & 0.489 & 0.571 & 0.601 & 0.503 & 0.529 & 0.582 & 0.483 & 0.852 \\
& N / N & 75 & 0.481 & 3.50 & 3.05 & 2.38 & \textbf{3.08} & 0.481 & 0.484 & 0.541 & 0.636 & 0.501 & 0.515 & 0.625 & 0.515 & 0.821 \\
\bottomrule
\end{tabular}
}
\caption{Consolidated evaluation results of taxonomies constructed with \textsc{TaxonomyBuilder} across the NLx and USAJOBS Datasets. The different configurations are represented by whether they include data \textbf{aug}mentation (Y) or not (N), and similarly for \textbf{soft} clustering or not. LLM-as-a-Judge categories represent the average of their component metrics. The \textit{Best Util} column reports the Label Utilization corresponding to the threshold ($\tau$) that achieved the highest \textit{strict} coverage. \textbf{Bold} values indicate the best score in each column per dataset.}
\label{tab:results_consolidated}
\end{table*}

\paragraph{LLM-as-a-Judge.}
To evaluate a constructed taxonomy holistically, we adopt the LLM-as-a-Judge approach of \citet{zhang-etal-2025-llmtaxo}, who design an evaluation based on taxonomy metrics from \citet{10.1145/3530019.3535305}, for a set of four evaluation categories:

\begin{itemize}[leftmargin=0.85em]
    \itemsep -0.3em
    \item \textbf{Clarity}: consists of four metrics, namely \textit{precision}, \textit{ambiguity}, \textit{consistency}, and \textit{accessibility}.
    \item \textbf{Hierarchical Coherence}: consists of \textit{gradational specificity} and \textit{parent-child coherence}.
    \item \textbf{Orthogonality}: \textit{distinctiveness} and \textit{non-overlap}.
    \item \textbf{Completeness}: \textit{coverage}, \textit{depth}, and \textit{balance}.
\end{itemize}

These metrics are defined in the LLM-as-a-Judge prompt we use, which is slightly adapted from \citet{zhang-etal-2025-llmtaxo}. This is found in Table \ref{tab:prompt_judge} of the Appendix. We use \textsc{gpt-5.4-nano} (2026-03-17) \cite{singh2025openaigpt5card} with a temperature of 0.

\paragraph{Domain coverage.}
To evaluate the generalizability and representativeness of a constructed taxonomy, we perform domain coverage tests, adapted from the evaluation of \citet{li-etal-2025-building}.

This metric measures how well the base level statements (``labels'') of a taxonomy can be used to map and classify real-world data. We first prepare a test set from the held-out data from each dataset (see Section \ref{sec:datasets}), namely a random sample 10k job postings from NLx and 1k postings from USAJOBS. We sentence tokenize all of these postings (using \textsc{NLTK}), and prompt \textsc{gpt-5.4-nano} and Google's \textsc{gemini-3.1-flash-lite} (with temperature of 0) to label each sentence as an AI-related job skill or not. The full prompt used for this is found in Table \ref{tab:prompt_label} of the Appendix.

With the labeled test sets, we evaluate coverage by running embedding-based semantic search procedures at various \textit{confidence levels}, or thresholds. We choose $\tau \in \{0.9, 0.8, 0.7, 0.6\}$ indicating that a sentence in question must match a taxonomy label with cosine similarity $> \tau$ to be classified as ``AI''. To calculate the similarity score for thresholding, we use an ensemble of three embedding models, akin to ensemble used for class scoring in Section \ref{sec:method}. At each $\tau$, we report the macro-F1 score, which represents the \textit{coverage rate}. This includes a \textit{lenient} score, where the ground truth of ``AI'' (label = 1) occurs when at least one LLM label is 1 (see above), and \textit{strict} where both LLMs agree. At the best (strict) coverage rate, we also report the \textit{label utilization rate}, i.e., the percentage of taxonomy statements successfully matched to the test set in the optimal (highest coverage) scenario. Together, these metrics illustrate both \textit{how well} a taxonomy represents a domain and \textit{how efficiently} this is done with respect to the breadth of the taxonomy.

\begin{table*}[htbp]
\centering
\resizebox{0.97\textwidth}{!}{
\begin{tabular}{ll l ccc l}
\toprule
\textbf{Dataset} & \textbf{Metric} & \textbf{Significant Factor} & \textbf{$F$-value} & \textbf{$p$-value} & \textbf{$\eta^2$} & \textbf{Direction of Effect / Explanation} \\
\midrule

\multirow{6}{*}{\textbf{NLx}} 
& \multirow{2}{*}{Silhouette Score} & Percentile & 4.36 & 0.059$^\dagger$ & 0.387 & Stricter percentile filtering improves cluster fit \\
& & Augmentation & 3.68 & 0.097$^\dagger$ & 0.163 & Augmentation marginally decreases cluster fit \\
\cmidrule(lr){2-7}
& LLM: Orthogonality & Percentile & 3.45 & 0.091$^\dagger$ & 0.467 & Stricter percentile filtering improves orthogonality \\
\cmidrule(lr){2-7}
& LLM: Completeness & Soft Clustering & 3.92 & 0.088$^\dagger$ & 0.223 & Soft clustering marginally improves completeness \\
\cmidrule(lr){2-7}
& Coverage (Strict, $\tau = 0.8$) & Augmentation & 40.81 & \textbf{<0.001}* & 0.836 & Augmentation severely reduces strict coverage rate \\
\cmidrule(lr){2-7}
& Label Utilization & Percentile & 4.60 & 0.053$^\dagger$ & 0.475 & Stricter filtering drastically improves label utilization \\
\midrule
\multirow{2}{*}{\textbf{USAJOBS}}
& Silhouette Score & Soft Clustering & 18.76 & \textbf{0.003}* & 0.713 & Soft clustering severely degrades cluster fit \\
\cmidrule(lr){2-7}
& Label Utilization & Augmentation & 13.48 & \textbf{0.008}* & 0.518 & Augmentation severely reduces label utilization \\
\bottomrule
\multicolumn{7}{l}{\footnotesize * Significant at $p < 0.05$; $^\dagger$ Marginally significant at $p < 0.10$.}
\end{tabular}
}
\caption{Factorial ANOVA results for effects of taxonomy construction configurations on the reported metrics. Only significant ($p < 0.05$) and marginally significant ($p < 0.10$) effects are reported. Effect sizes are denoted by $\eta^2$.}
\label{tab:anova_sig}
\end{table*}

\section{Results and Statistical Analysis}
The consolidated results are presented in Table \ref{tab:results_consolidated}, where LLM-as-a-Judge scores are reported as averages across the four categories. The label utilization rate is only reported for the corresponding best strict coverage rate. For the complete results, we refer the readers to Table \ref{tab:results_full} of the Appendix.

An analysis of the results shows that the best results are achieved generally when \textit{either} data augmentation \textit{or} soft clustering is used, and a majority (22/28) of the best scores per metric are achieved at or above the 50th percentile filtering mark. To verify these observations, as well as to investigate the main effects of the different taxonomy construction configurations, we perform a macro-level statistical analysis using a factorial ANOVA test. The ANOVA test was conducted in the aggregated metrics (i.e., as presented in Table \ref{tab:results_full}) to evaluate the effects of data augmentation, percentile filtering, and soft clustering on our evaluation metrics (2 $\times$ 3 $\times$ 2 factors). The statistically significant results of this analysis are presented in Table \ref{tab:anova_sig}, and the full results can be found in Table \ref{tab:anova_full} of the Appendix.

\section{Discussion}
In the following, we reflect on the main findings of our work, and we discuss paths for future research.

\paragraph{Less is more.}
Our work centers on the important question of systemically understanding complex, shifting, and evolving domains such as \textit{AI skills}, specifically through the construction of representative taxonomies from massive text corpora. While the presence of such data may potentially serve as a fertile foundation for the challenging task of automatic taxonomy construction, we demonstrate that there are important design decisions revolving around \textit{how to leverage this data}. Such questions have not been explored by previous works, who operate on much more compact datasets.

Several of our key findings point to a \textit{less is more} philosophy when building taxonomies from large-scale data. We find that data augmentation does not help the quality of the resulting taxonomies, demonstrated by the fact that only one LLM-as-a-Judge category was won with data augmentation (out of the eight settings). In this light, using soft clustering after \textit{excluding} data augmentation proved to be a winning recipe, achieving the best score in 20/28 cases. Thus, soft clustering likely only makes sense when not including augmented samples.

Studying the effect of percentile-based filtering measure reveals a similar trend. Taxonomies built with a weaker 25th percentile filter won zero LLM-as-a-Judge categories, and interestingly, only outperform all other taxonomies at very strict (high $\tau$) coverage tests. As such, being too lenient with data inclusion (i.e., including more noisy points), can lead to sub-optimal domain-specific taxonomies.

\paragraph{Taxonomy building as trade-off balancing.}
We highlight another important finding, specifically in evaluating the \textit{coverage} of taxonomies in a custom domain. One can observe an inverse relationship between data inclusion and coverage performance. Interestingly, at the strictest coverage threshold ($\tau = 0.9$), the coverage rates (both strict and lenient) are won exclusively by taxonomies built with a 25th percentile filter (more data). This is exactly the opposite for the lowest threshold ($\tau = 0.6$), at which exclusively 75th percentile taxonomies outperform all others. Intuitively, this might make sense, as larger, potentially noisy taxonomies may provide wide coverage and thereby excel at higher similarity thresholds, whereas more concise taxonomies can still perform better at lower thresholds due to being more well-defined and less cluttered.

Such findings point to the need to study \textit{trade-offs} in automatic taxonomy construction, particularly in the balance between domain coverage and intra-taxonomy variability (i.e., noisiness). Data augmentation and more permissive filtering could serve certain use cases well, in which breadth is prioritized over intepretability. On the other hand, taxonomies that are repetitive or not focused enough on the domain in question may only increase ambiguity and lack usability or complicate comprehension.  Therefore, although our results support the recommendation of less (data) being more powerful, we caution that this is case-specific, and ultimately, trade-offs should be carefully weighed.

\section{Conclusion}
We propose \textsc{TaxonomyBuilder} as a generalizable tool to taxonomize custom domains from large-scale text corpora. Using \textsc{TaxonomyBuilder}, we conduct a systematic investigation into the practical considerations of leveraging large amounts of data, focusing on questions of optimal data use for grounding taxonomy construction. Our results highlight that less is more, and data augmentation should be exercised with caution. In addition, performing selective filtering of candidate inputs to the construction process is often advantageous. As future work, we recommend (1) the continued exploration of design decisions in automatic taxonomy construction, (2) the carrying out of ablation studies regarding LLM choice, clustering parametrization, and taxonomy pruning, and (3) the formalization of trade-off-specific taxonomy evaluation metrics, focusing on real-world generalizability.

\newpage
\section*{Limitations}
We acknowledge the main limitations of our work, primarily in that we did not study the impact our our investigated factors (data augmentation, soft clustering, and percentile filtering) on other proposed LLM-based taxonomy construction methods. Rather, we focused on our proposed \textsc{TaxonomyBuilder} method, as this framework was custom built for this systematic study, thereby not necessitating the modification of previous methods.

We also note the lack of an ablation study, as mentioned in the Conclusion, which we propose as direct follow-up work to ours. Important points of investigation could include but are not limited to: choice of LLM for label generation, the design of more intelligent methods to reduce noise in the context provided to the LLM (while maintaining representativeness, and conducting top-down checks to counteract bottom-up error propagation (e.g., induced via irrelevant candidate sentences).

\section*{Ethics Statement}
We respect the confidentiality of data shared under the agreement with NLx and publicly release only aggregate information (the constructed taxonomies), and mitigate restrictions on sharing with the inclusion of a public corpus in the same domain.

We also note that the constructed taxonomies provided in our repository have been redacted to remove company names remaining in the leaf nodes.


\bibliography{custom}

\appendix

\section{Supplemental}
\label{sec:appendix}

\paragraph{Prompts.}
Table \ref{tab:prompt_base} provides the LLM prompt used to create labels for collections of candidate sentences, in order to form a level of the taxonomy being constructed. Table \ref{tab:prompt_agg} provides the prompt for the aggregation process (redundancy reduction). 

On the evaluation side, Table \ref{tab:prompt_judge} outlines the full LLM-as-a-Judge prompt, based on \citet{zhang-etal-2025-llmtaxo}. Table \ref{tab:prompt_label} contains the prompt used to label the test sets for NLx and USAJOBS.

\paragraph{Results.}
Table \ref{tab:results_full} details the complete experimental results, providing a more full account than Table \ref{tab:results_consolidated}. Table \ref{tab:anova_full} presents the full ANOVA test results, in supplement to the abridged version of Table \ref{tab:anova_sig}.

\begin{table}[htbp]
\centering
\scriptsize
\begin{tabular}{p{0.95\linewidth}}
\hline
You will be given a list of statements. \\
Your job is to produce a single sentence that summarizes these statements into a coherent task / skill description. \\
A task is a specfic activity or function that a person would be required to do on the job. \\
A skill is the ability to perform a specific task and apply knowledge, particularly in the work context. \\
Avoid using generalizations like "various" and "across domains". \\
Answer simply with the generated description, nothing else is required. \\\\

Provide your feedback as follows:\\\\

Output:::\\
Description: (GENERATED DESCRIPTION)\\\\

Here are some examples:\\

statements: ['develop or implement data analysis algorithms.',
  'design and apply bioinformatics algorithms including unsupervised and supervised machine learning, dynamic programming, or graphic algorithms.',
  'analyze or manipulate bioinformatics data using software packages, statistical applications, or data mining techniques.',
  'develop or apply data mining and machine learning algorithms.',
  'develop machine learning operations.',
  '+ develop and implement new computational and statistical methods.',
  'he/she may also evaluate and develop novel algorithms and approaches for data analysis.',
  '* you will develop custom data models and algorithms to apply to data sets.'] \\\\

Output:::\\
Description: Develop or apply data mining and machine learning algorithms.\\\\

statements: ['• basic computer skills.',
  'basic computer software skills.',
  'basic computer software skills, i.e.',
  'be able to perform basic computer skills.',
  'basic skills in use of computers and software programs.',
  'skills: basic computer skills.',
  '* basic programming experience.',
  'computer skills and basic knowledge of software applications.',
  '•basic computer skills.',
  '- basic computer skills.',
  'skills: basic computer knowledge.',
  'ability to perform basic computer skills.',
  'basic problem solving skills associated with software applications used is expected.',
  '- computer programming skills.']\\\\

Output:::\\
Description: Has basic computer skills, including programming and use of computers and software programs.\\\\

Now here are the actual statements.\\\\

statements: \{CANDIDATES\}\\\\

Output:::\\
Description:\\
\hline
\end{tabular}
\caption{Prompt for creation of leaf-level taxonomy entries.}
\label{tab:prompt_base}
\end{table}

\begin{table}[htbp]
\centering
\small
\begin{tabular}{p{0.95\linewidth}}
\hline
You will be given a list of statements.\\
They all express a similar meaning, with slight variations and differences.\\
Your job is to produce a single sentence that summarizes these statements into one coherent description, that captures the essence and intricacies of all statements.\\
Answer simply with the generated description, nothing else is required.\\\\

Provide your feedback as follows:\\\\

Output:::\\
Description: (GENERATED DESCRIPTION)\\\\

Now here are the actual statements.\\\\

statements: \{LABELS\}\\\\

Output:::\\\\
Description: \\
\hline
\end{tabular}
\caption{Prompt for aggregation of similar labels (for label pruning).}
\label{tab:prompt_agg}
\end{table}

\begin{table}[t]
\centering
\scriptsize
\begin{tabular}{p{0.95\linewidth}}
\hline
You will be given a taxonomy related to AI skills in the workplace. Please evaluate the taxonomy in the json format using the evaluation metrics provided. Give each evaluation criteria a score from 1-5. No additional explanation is necessary. Output only the json results.\\
        *********************************\\
            Below is the taxonomy:\\
            \{TAXONOMY JSON STRING\}\\

    *******************************\\
    Here are the metrics:\\
    Taxonomy Evaluation Metrics\\
Give the score from [1-5] 5: best/strongly agree, 1: worst/strongly disagree\\ \\

Clarity: Assess whether the topic labels are clear, precise, and unambiguous. \\Purpose: Ensure that each topic label communicates its content effectively to avoid confusion. \\Evaluation Criteria:\\
•	Precision: Each topic label uses specific and well-defined terms.\\
o	Score: \\
•	Unambiguity: Topic labels should have only one interpretation, preventing misunderstanding.\\
o	Score: \\
•	Consistency: Use of terminology is consistent across all levels of the taxonomy.\\
o	Score:\\
•	Accessibility: Language is straightforward, avoiding jargon where possible unless it is standard within the covered domain.\\
o	Score:\\\\

Hierarchical Coherence:  Assess whether the taxonomy follows a clear and meaningful hierarchical structure. \\Purpose: Ensure that the taxonomy’s structure facilitates easy navigation and understanding by clearly organizing information from the most general to the most specific. \\Evaluation Criteria:\\
•	Gradational Specificity: There is a logical progression from broader to more specific categories.\\
o	Score:\\
•	Parent-Child Coherence: Parent-child relationships are well-formed, ensuring that child nodes logically belong to their parent nodes.\\
o	Score:\\
•	Consistency: The hierarchy maintains consistent levels of detail throughout the taxonomy, ensuring that no topics are too broad or too narrow relative to others at the same level.\\
o	Score:\\\\

Orthogonality: Assess whether the topics are well-differentiated without duplication. \\Purpose: Maintain distinct boundaries between topics to ensure that each topic captures unique aspects of the domain. \\Evaluation Criteria:\\
•	Distinctiveness: Topics at each level progressively add meaningful distinctions rather than just rephrasing broader topics.\\
o	Score:\\
•	Non-overlap: For each topic, there is minimal to no overlap in the scope or content with other topics.\\
o	Score:\\\\

Completeness: Assess whether the taxonomy captures a broad and representative set of topics across different aspects of the domain. \\Purpose: Cover as many areas of the topic to ensure the taxonomy is comprehensive. \\Evaluation Criteria:\\
•	Domain Coverage: The taxonomy covers a variety of significant aspects of the domain it represents.\\
o	Score:\\
•	Depth: The taxonomy provides sufficient depth in each branch to capture nuanced distinctions within topics.\\
o	Score:\\
•	Balance: The topics are evenly distributed across the taxonomy. This involves assessing whether some branches are disproportionately detailed while others are underdeveloped, which could lead to an imbalance that might skew the taxonomy’s effectiveness and navigability.\\
o	Score:\\\\

    Important: ensure that the output is in json, and following the above-given criteria names exactly.\\\\

    Now please provide the scores.\\
\hline
\end{tabular}
\caption{LLM-as-a-Judge prompt for comprehensive taxonomy evaluation.}
\label{tab:prompt_judge}
\end{table}

\begin{table}[htbp]
\centering
\small
\begin{tabular}{p{0.95\linewidth}}
\hline
You will be given a dictionary of statements from a job posting, each mapped to a unique ID (integer).\\
Your job is to indicate which of the statements (marked by their IDs), contain information related to any Artificial Intelligence (AI) task, skill, activity, expertise, or requirement.\\
"AI-related" can also include soft skills required for AI jobs in addition to technical abilities or tasks.\\
Make sure, though, that only AI-related statements are chosen!\\
Answer simply with a comma-separated Python list of ONLY the statement IDs marked as AI-related. Make sure to stick to this output format!\\\\

Provide your feedback as follows:\\\\

Output:::\\
Classification: (LIST OF IDENTIFIED AI-RELATED STATEMENT IDs)\\\\

Now here are the actual statements.\\\\

texts: \{DICTIONARY OF MAPPED JOB POSTING SENTENCES\}\\\\

Output:::\\
Classification:\\
\hline
\end{tabular}
\caption{Prompt for LLM labeling of test set for domain coverage tests.}
\label{tab:prompt_label}
\end{table}

\begin{table*}[htbp]
\centering
\resizebox{\textwidth}{!}{
\begin{tabular}{lll c cccc cc cc ccc ccc ccc ccc ccc}
\toprule
\multirow{2}{*}{\textbf{Data}} & \multicolumn{2}{c}{\textbf{Config}} & \multirow{2}{*}{\textbf{Silh.}} & \multicolumn{4}{c}{\textbf{Clarity}} & \multicolumn{2}{c}{\textbf{H. Coh.}} & \multicolumn{2}{c}{\textbf{Orth.}} & \multicolumn{3}{c}{\textbf{Comp.}} & \multicolumn{3}{c}{\textbf{Cov (0.9)}} & \multicolumn{3}{c}{\textbf{Cov (0.8)}} & \multicolumn{3}{c}{\textbf{Cov (0.7)}} & \multicolumn{3}{c}{\textbf{Cov (0.6)}} \\
\cmidrule(lr){2-3} \cmidrule(lr){5-8} \cmidrule(lr){9-10} \cmidrule(lr){11-12} \cmidrule(lr){13-15} \cmidrule(lr){16-18} \cmidrule(lr){19-21} \cmidrule(lr){22-24} \cmidrule(lr){25-27}
& \textbf{Aug/Soft} & \textbf{Pct} & & \textbf{Pr} & \textbf{Un} & \textbf{Co} & \textbf{Ac} & \textbf{Gr} & \textbf{PC} & \textbf{Di} & \textbf{NO} & \textbf{Co} & \textbf{De} & \textbf{Ba} & \textbf{L} & \textbf{S} & \textbf{U} & \textbf{L} & \textbf{S} & \textbf{U} & \textbf{L} & \textbf{S} & \textbf{U} & \textbf{L} & \textbf{S} & \textbf{U} \\
\midrule
\multirow{12}{*}{\textbf{NLx}} 
& Y / Y & 25 & 0.047 & 3.14 & 2.33 & 2.14 & 2.86 & 2.86 & 3.29 & 2.10 & 1.14 & 3.62 & 3.43 & 2.00 & 0.472 & 0.498 & 0.101 & 0.595 & 0.545 & 0.580 & 0.585 & 0.447 & 0.813 & 0.346 & 0.227 & 0.847 \\
& Y / Y & 50 & 0.056 & 3.50 & 2.69 & 2.13 & 3.13 & 3.06 & 3.44 & 2.19 & 1.44 & 3.75 & 3.13 & 2.00 & 0.470 & 0.497 & 0.091 & 0.592 & 0.547 & 0.604 & 0.603 & 0.464 & 0.838 & 0.376 & 0.253 & 0.871 \\
& Y / Y & 75 & 0.092 & 3.71 & 2.86 & 2.43 & 3.29 & 3.43 & 3.71 & 2.57 & 1.86 & 4.00 & 3.71 & 2.14 & 0.462 & 0.494 & 0.058 & 0.579 & 0.557 & 0.610 & 0.630 & 0.494 & 0.867 & 0.427 & 0.297 & 0.899 \\
\cmidrule(lr){2-27}
& Y / N & 25 & -0.000 & 3.43 & 2.43 & 2.14 & 3.57 & 2.86 & 3.29 & 2.29 & 1.29 & 3.71 & 3.00 & 2.00 & 0.475 & 0.500 & 0.113 & 0.592 & 0.540 & 0.580 & 0.576 & 0.436 & 0.815 & 0.316 & 0.201 & 0.852 \\
& Y / N & 50 & 0.032 & 3.50 & 2.50 & 2.20 & 3.10 & 2.90 & 3.40 & 2.30 & 1.40 & 3.60 & 3.20 & 2.00 & 0.473 & 0.499 & 0.104 & 0.590 & 0.543 & 0.608 & 0.597 & 0.458 & 0.850 & 0.362 & 0.241 & 0.881 \\
& Y / N & 75 & 0.074 & 3.57 & 2.71 & 2.00 & 3.00 & 3.00 & 3.71 & 2.57 & 1.71 & 3.57 & 3.14 & 2.00 & 0.465 & 0.495 & 0.067 & 0.576 & 0.553 & 0.604 & 0.626 & 0.488 & 0.870 & 0.402 & 0.276 & 0.907 \\
\cmidrule(lr){2-27}
& N / Y & 25 & 0.073 & 3.40 & 2.40 & 2.00 & 3.20 & 2.80 & 3.60 & 2.40 & 1.40 & 3.80 & 3.40 & 2.00 & 0.460 & 0.493 & 0.042 & 0.554 & 0.600 & 0.468 & 0.702 & 0.599 & 0.824 & 0.546 & 0.398 & 0.897 \\
& N / Y & 50 & 0.111 & 4.00 & 3.00 & 2.83 & 3.17 & 3.50 & 4.00 & 2.67 & 1.83 & 4.00 & 3.67 & 2.00 & 0.457 & 0.489 & 0.031 & 0.538 & 0.598 & 0.460 & 0.707 & 0.636 & 0.814 & 0.602 & 0.445 & 0.908 \\
& N / Y & 75 & 0.065 & 3.50 & 2.50 & 2.50 & 3.00 & 3.00 & 3.50 & 2.50 & 1.50 & 4.00 & 3.50 & 2.00 & 0.456 & 0.489 & 0.025 & 0.504 & 0.570 & 0.447 & 0.682 & 0.672 & 0.844 & 0.663 & 0.505 & 0.936 \\
\cmidrule(lr){2-27}
& N / N & 25 & 0.026 & 3.67 & 2.67 & 2.00 & 3.33 & 3.00 & 3.67 & 2.33 & 1.33 & 3.67 & 3.00 & 2.00 & 0.461 & 0.497 & 0.056 & 0.558 & 0.596 & 0.447 & 0.700 & 0.589 & 0.829 & 0.533 & 0.385 & 0.916 \\
& N / N & 50 & 0.079 & 3.67 & 2.67 & 2.00 & 3.00 & 3.33 & 3.67 & 2.33 & 1.33 & 4.00 & 3.67 & 2.00 & 0.459 & 0.493 & 0.043 & 0.541 & 0.596 & 0.447 & 0.705 & 0.623 & 0.837 & 0.599 & 0.441 & 0.924 \\
& N / N & 75 & 0.096 & 3.67 & 2.67 & 2.33 & 3.00 & 3.33 & 3.67 & 2.33 & 1.67 & 4.00 & 3.33 & 2.00 & 0.456 & 0.489 & 0.031 & 0.510 & 0.576 & 0.451 & 0.685 & 0.662 & 0.851 & 0.648 & 0.488 & 0.943 \\
\midrule
\multirow{12}{*}{\textbf{USAJOBS}} 
& Y / Y & 25 & 0.155 & 3.92 & 3.45 & 2.61 & 3.37 & 2.16 & 3.24 & 2.45 & 1.83 & 3.02 & 2.55 & 2.63 & 0.485 & 0.510 & 0.060 & 0.514 & 0.518 & 0.294 & 0.543 & 0.469 & 0.846 & 0.319 & 0.244 & 0.931 \\
& Y / Y & 50 & 0.159 & 3.97 & 3.54 & 2.75 & 3.33 & 2.18 & 3.43 & 2.38 & 1.89 & 3.11 & 2.78 & 2.64 & 0.483 & 0.506 & 0.070 & 0.496 & 0.526 & 0.300 & 0.573 & 0.518 & 0.848 & 0.436 & 0.354 & 0.933 \\
& Y / Y & 75 & 0.430 & 3.50 & 2.50 & 2.00 & 3.00 & 2.00 & 1.50 & 2.00 & 1.00 & 3.00 & 2.00 & 2.00 & 0.481 & 0.501 & 0.035 & 0.490 & 0.525 & 0.253 & 0.586 & 0.559 & 0.823 & 0.584 & 0.466 & 0.924 \\
\cmidrule(lr){2-27}
& Y / N & 25 & 0.561 & 3.90 & 3.51 & 2.73 & 3.34 & 2.19 & 3.32 & 2.57 & 2.00 & 3.01 & 2.59 & 2.73 & 0.486 & 0.507 & 0.066 & 0.514 & 0.518 & 0.296 & 0.553 & 0.476 & 0.835 & 0.330 & 0.254 & 0.922 \\
& Y / N & 50 & 0.516 & 3.96 & 3.62 & 2.87 & 3.44 & 2.16 & 3.41 & 2.50 & 2.03 & 3.09 & 2.84 & 2.91 & 0.483 & 0.506 & 0.070 & 0.498 & 0.525 & 0.317 & 0.576 & 0.517 & 0.855 & 0.425 & 0.345 & 0.948 \\
& Y / N & 75 & 0.577 & 3.33 & 2.67 & 2.67 & 3.67 & 2.33 & 2.67 & 2.00 & 1.00 & 3.00 & 2.00 & 2.67 & 0.481 & 0.500 & 0.028 & 0.491 & 0.531 & 0.271 & 0.583 & 0.552 & 0.839 & 0.556 & 0.446 & 0.947 \\
\cmidrule(lr){2-27}
& N / Y & 25 & 0.345 & 3.67 & 3.44 & 1.89 & 4.33 & 1.56 & 1.67 & 1.44 & 1.67 & 1.67 & 1.78 & 2.67 & 0.483 & 0.505 & 0.053 & 0.492 & 0.534 & 0.275 & 0.587 & 0.572 & 0.872 & 0.587 & 0.469 & 0.947 \\
& N / Y & 50 & 0.362 & 3.57 & 3.71 & 2.29 & 4.43 & 1.57 & 2.29 & 1.71 & 2.43 & 1.86 & 1.71 & 2.57 & 0.481 & 0.503 & 0.028 & 0.489 & 0.527 & 0.245 & 0.577 & 0.588 & 0.862 & 0.607 & 0.486 & 0.946 \\
& N / Y & 75 & 0.112 & 4.26 & 3.85 & 3.07 & 3.30 & 2.67 & 3.93 & 2.85 & 2.07 & 3.33 & 3.00 & 2.85 & 0.480 & 0.499 & 0.014 & 0.484 & 0.516 & 0.157 & 0.542 & 0.637 & 0.829 & 0.644 & 0.526 & 0.968 \\
\cmidrule(lr){2-27}
& N / N & 25 & 0.673 & 3.42 & 3.42 & 2.33 & 4.17 & 1.58 & 1.83 & 1.50 & 1.83 & 1.67 & 1.67 & 2.58 & 0.482 & 0.506 & 0.050 & 0.491 & 0.533 & 0.255 & 0.586 & 0.571 & 0.835 & 0.589 & 0.470 & 0.950 \\
& N / N & 50 & 0.703 & 2.50 & 2.00 & 2.00 & 3.50 & 1.50 & 1.50 & 2.00 & 1.00 & 3.00 & 2.50 & 1.50 & 0.482 & 0.503 & 0.056 & 0.489 & 0.529 & 0.237 & 0.571 & 0.582 & 0.852 & 0.601 & 0.483 & 0.959 \\
& N / N & 75 & 0.481 & 4.20 & 3.70 & 2.90 & 3.20 & 2.40 & 3.70 & 2.75 & 2.00 & 3.50 & 3.05 & 2.70 & 0.481 & 0.501 & 0.037 & 0.484 & 0.515 & 0.174 & 0.541 & 0.625 & 0.821 & 0.636 & 0.515 & 0.972 \\
\bottomrule
\end{tabular}
}
\caption{Comprehensive evaluation results of taxonomies constructed with \textsc{TaxonomyBuilder} across the NLx and USAJOBS Datasets. The different configurations are represented by whether they include data augmentation (Y) or not (N), and similarly for soft clustering or not. \textit{Pct} denotes the percentile filtering threshold. All LLM-as-a-Judge metrics are reported in the order they were introduced in the paper, and are abbreviated with two representative letters. Domain Coverage (\textit{Cov)} metrics include Lenient (L), Strict (S), and Utilization (U) across four thresholds.}
\label{tab:results_full}
\end{table*}

\begin{table}[htbp]
\centering
\small
\resizebox{0.99\linewidth}{!}{
\begin{tabular}{ll l r r r}
\toprule
\textbf{Dataset} & \textbf{Evaluation Metric} & \textbf{Factor} & \textbf{$F$-value} & \textbf{$p$-value} & \textbf{$\eta^2$} \\
\midrule
\multirow{21}{*}{\textbf{NLx}} 
& \multirow{3}{*}{Silhouette Score} 
  & Augmentation & 3.68 & 0.097$^\dagger$ & 0.163 \\
& & Soft Clustering & 3.11 & 0.121\phantom{$^\dagger$} & 0.138 \\
& & Percentile & 4.36 & 0.059$^\dagger$ & 0.387 \\
\cmidrule(lr){2-6}
& \multirow{3}{*}{LLM: Clarity} 
  & Augmentation & 0.63 & 0.454\phantom{$^\dagger$} & 0.065 \\
& & Soft Clustering & 0.14 & 0.716\phantom{$^\dagger$} & 0.015 \\
& & Percentile & 0.91 & 0.444\phantom{$^\dagger$} & 0.190 \\
\cmidrule(lr){2-6}
& \multirow{3}{*}{LLM: Hierarchical Coherence} 
  & Augmentation & 2.82 & 0.137\phantom{$^\dagger$} & 0.190 \\
& & Soft Clustering & 0.09 & 0.770\phantom{$^\dagger$} & 0.006 \\
& & Percentile & 2.48 & 0.153\phantom{$^\dagger$} & 0.334 \\
\cmidrule(lr){2-6}
& \multirow{3}{*}{LLM: Orthogonality} 
  & Augmentation & 0.48 & 0.512\phantom{$^\dagger$} & 0.032 \\
& & Soft Clustering & 0.38 & 0.555\phantom{$^\dagger$} & 0.026 \\
& & Percentile & 3.45 & 0.091$^\dagger$ & 0.467 \\
\cmidrule(lr){2-6}
& \multirow{3}{*}{LLM: Completeness} 
  & Augmentation & 2.98 & 0.128\phantom{$^\dagger$} & 0.170 \\
& & Soft Clustering & 3.92 & 0.088$^\dagger$ & 0.223 \\
& & Percentile & 1.84 & 0.228\phantom{$^\dagger$} & 0.209 \\
\cmidrule(lr){2-6}
& \multirow{3}{*}{Strict Coverage ($\tau=0.8$)} 
  & Augmentation & 40.81 & \textbf{<0.001}* & 0.836 \\
& & Soft Clustering & 0.11 & 0.750\phantom{$^\dagger$} & 0.002 \\
& & Percentile & 0.46 & 0.650\phantom{$^\dagger$} & 0.019 \\
\cmidrule(lr){2-6}
& \multirow{3}{*}{Best Label Utilization} 
  & Augmentation & 3.18 & 0.118\phantom{$^\dagger$} & 0.164 \\
& & Soft Clustering & <0.01 & 0.986\phantom{$^\dagger$} & <0.001 \\
& & Percentile & 4.60 & 0.053$^\dagger$ & 0.475 \\
\midrule
\multirow{21}{*}{\textbf{USAJOBS}}
& \multirow{3}{*}{Silhouette Score} 
  & Augmentation & 0.38 & 0.556\phantom{$^\dagger$} & 0.015 \\
& & Soft Clustering & 18.76 & \textbf{0.003}* & 0.713 \\
& & Percentile & 0.09 & 0.912\phantom{$^\dagger$} & 0.007 \\
\cmidrule(lr){2-6}
& \multirow{3}{*}{LLM: Clarity} 
  & Augmentation & 0.07 & 0.797\phantom{$^\dagger$} & 0.009 \\
& & Soft Clustering & 0.25 & 0.634\phantom{$^\dagger$} & 0.033 \\
& & Percentile & 0.11 & 0.895\phantom{$^\dagger$} & 0.030 \\
\cmidrule(lr){2-6}
& \multirow{3}{*}{LLM: Hierarchical Coherence} 
  & Augmentation & 0.83 & 0.392\phantom{$^\dagger$} & 0.094 \\
& & Soft Clustering & 0.01 & 0.933\phantom{$^\dagger$} & 0.001 \\
& & Percentile & 0.50 & 0.628\phantom{$^\dagger$} & 0.113 \\
\cmidrule(lr){2-6}
& \multirow{3}{*}{LLM: Orthogonality} 
  & Augmentation & 0.01 & 0.920\phantom{$^\dagger$} & 0.002 \\
& & Soft Clustering & 0.03 & 0.873\phantom{$^\dagger$} & 0.004 \\
& & Percentile & 0.03 & 0.973\phantom{$^\dagger$} & 0.008 \\
\cmidrule(lr){2-6}
& \multirow{3}{*}{LLM: Completeness} 
  & Augmentation & 1.27 & 0.297\phantom{$^\dagger$} & 0.127 \\
& & Soft Clustering & 0.18 & 0.686\phantom{$^\dagger$} & 0.018 \\
& & Percentile & 0.79 & 0.489\phantom{$^\dagger$} & 0.158 \\
\cmidrule(lr){2-6}
& \multirow{3}{*}{Strict Coverage ($\tau=0.8$)} 
  & Augmentation & 0.17 & 0.691\phantom{$^\dagger$} & 0.021 \\
& & Soft Clustering & 0.04 & 0.856\phantom{$^\dagger$} & 0.004 \\
& & Percentile & 0.48 & 0.640\phantom{$^\dagger$} & 0.117 \\
\cmidrule(lr){2-6}
& \multirow{3}{*}{Best Label Utilization} 
  & Augmentation & 13.48 & \textbf{0.008}* & 0.518 \\
& & Soft Clustering & <0.01 & 0.974\phantom{$^\dagger$} & <0.001 \\
& & Percentile & 2.77 & 0.130\phantom{$^\dagger$} & 0.213 \\
\bottomrule
\multicolumn{6}{l}{\footnotesize * Significant at $p < 0.05$; $^\dagger$ Marginally significant at $p < 0.10$.}
\end{tabular}
}
\caption{Complete factorial ANOVA results. Significant ($p < 0.05$) and marginally significant ($p < 0.10$) effects are denoted. Effect size is represented by $\eta^2$.}
\label{tab:anova_full}
\end{table}

\end{document}